\documentclass{article} 
\usepackage[numbers,sort&compress]{natbib}
\usepackage[dblblindworkshop,final]{neurips_2026}
\usepackage{times}

\usepackage{amsmath,amsfonts,bm}

\def\eqref#1{equation~\ref{#1}}

\def\1{\bm{1}}

\DeclareMathAlphabet{\mathsfit}{\encodingdefault}{\sfdefault}{m}{sl}
\SetMathAlphabet{\mathsfit}{bold}{\encodingdefault}{\sfdefault}{bx}{n}

\usepackage{hyperref}
\usepackage{url}

\usepackage{booktabs}
\usepackage{amsmath}
\usepackage{amssymb}
\usepackage{siunitx}
\usepackage{graphicx}
\usepackage{caption}
\usepackage{enumitem}
\usepackage{amsthm}

\usepackage{todonotes}

\usepackage[most]{tcolorbox}

\newtcolorbox{summary}[1][]{
  colback=gray!5,
  colframe=gray!50!black,
  fonttitle=\bfseries,
  title=Summary,
  #1
}

\newtheorem{theorem}{Theorem}
\theoremstyle{definition}
\newtheorem{definition}[theorem]{Definition}

\tcbset{thmbox/.style={
  enhanced, sharp corners, breakable,
  boxrule=0.4pt, colback=black!3, colframe=black!60,
  boxsep=0pt, left=3pt, right=3pt, top=2pt, bottom=2pt,
  before skip=4pt, after skip=4pt,
  enlarge left by=-3pt, enlarge right by=-3pt,
  width=\dimexpr\linewidth+6pt\relax,
}}

\newcommand{\tightdisplays}{%
  \setlength{\abovedisplayskip}{3pt}%
  \setlength{\belowdisplayskip}{1pt}%
  \setlength{\abovedisplayshortskip}{1pt}%
  \setlength{\belowdisplayshortskip}{1pt}}

\NewDocumentEnvironment{btheorem}{o}
  {\begin{tcolorbox}[thmbox]\tightdisplays
   \IfValueTF{#1}{\begin{theorem}[#1]}{\begin{theorem}}}
  {\end{theorem}\end{tcolorbox}}

\NewDocumentEnvironment{bdefinition}{o}
  {\begin{tcolorbox}[thmbox]\tightdisplays
   \IfValueTF{#1}{\begin{definition}[#1]}{\begin{definition}}}
  {\end{definition}\end{tcolorbox}}
  
\newcommand{\SOthree}{\ensuremath{\mathrm{SO}(3)}}
\newcommand{\SOtwo}{\ensuremath{\mathrm{SO}(2)}}
\newcommand{\Stwo}{\ensuremath{S^2}}
\newcommand{\SOthreeSOtwo}{\ensuremath{\mathrm{SO}(3)/\mathrm{SO}(2)}}

\title{SO(3)-RoPE for Spherical Transformers}
\workshoptitle{Representations for the Physical Sciences}

\author{%
    Christian Libner$^{1}$\kern16pt
    Chase van de Geijn$^{1}$\kern16pt
    Alexander S. Ecker$^{1,2}$\kern16pt
    Maurice Weiler$^{3}$\\[8pt]
$^1$Institute of Computer Science \& CIDAS,\, University of Göttingen\\[.5pt]
$^2$Max Planck Institute for Dynamics and Self-Organization,\, Göttingen\\[.5pt]
$^3$MIT CSAIL\\
}

\begin{document}

\maketitle

\vspace*{-5pt}

\begin{abstract}
\vspace*{-5pt}
Spherical data arise in many scientific applications. Often spherical transformers disregard the geometry of the underlying spherical domain, causing distortions and coordinate singularities near the poles.
We introduce \SOthree{}-RoPE, a relative positional embedding that incorporates spherical geometry into transformer attention through unitary \SOthree{} representations.
Our formulation is \SOthree{}-equivariant and compatible with FlashAttention,
retaining efficiency of vanilla transformers.
On shallow water dynamics prediction over a rotating sphere,
our SO3ViT outperforms an S2Transformer baseline with lower errors and reduced runtime.
\end{abstract}

\vspace*{-5pt}
\section{Introduction}
\vspace*{-1pt}

Transformers have become a standard architecture across machine learning \citep{vaswani2017attention,dosovitskiy2020image,radford2023robust, jumper2021highly}.
Unlike convolutional networks, pure attention operations are not informed about spatial locations of their input tokens.
Positional encodings provide this missing information.
A widely used approach for encoding relative positions in sequences and Euclidean spaces  are \emph{Rotary Positional Embeddings} (RoPE) \citep{su2024roformer,kogkalidis2024algebraic}.
We extend this idea to spherical transformers with \emph{SO(3)-RoPE},
a relative positional encoding that incorporates the geometry of spherical data into attention.

Our approach builds on the established line of work on spherical convolutional neural networks \citep{Cohen2018SphericalC,weiler2026equivariant}.
A common operation in spherical CNNs is to lift feature maps from $S^2$ to the rotation group $\mathrm{SO}(3)$, where features depend both on their spatial location and orientation.
We employ such lifting convolutions as patch-embeddings of dense spherical data to tokens on $\mathrm{SO}(3)$.

$\mathrm{SO}(3)$-RoPE incorporates the geometry of these tokens into self-attention,
analogously to how conventional RoPE incorporates Euclidean geometry.
Euclidean RoPE acts via complex phase modulation (or rotations of \mbox{$2\!\times\!2$}-blocks of features),
which are unitary translation group representations.
Analogously, $\mathrm{SO}(3)$-RoPE acts via unitary representations of the rotation group,
specifically via Wigner D-matrices $D^\ell(g)$, the irreducible representations of $\mathrm{SO}(3)$.

Because SO(3)-RoPE modifies queries and keys independently,
our method can take advantage of hardware-optimised attention kernels like FlashAttention \citep{dao2022flashattention} out of the box.
As a consequence, it preserves similar throughput, memory efficiency and scalability as vanilla transformers.

\begin{figure}[t]
    \centering
    \includegraphics[]{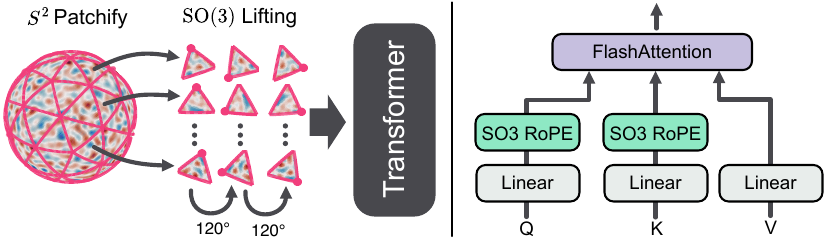}
    \caption{SO3ViT. We patchify the sphere and process those patches with a  transformer. In the attention we embed the queries and keys with our \SOthree{} RoPE.}
    \label{fig:overview}
\end{figure}

We evaluate our model SO3ViT (Fig. \ref{fig:overview})
on next-timestep prediction for shallow water dynamics on a rotating sphere.
Compared to an S2Transformer \citep{Bonev2025AttentionOT} baseline
--- a non-\SOthree{}-equivariant spherical transformer architecture ---
we are able to scale to larger models and achieve significantly lower prediction errors,
while drastically reducing runtime.
Moreover, S2Transformer suffers from high errors due to geometry distortion around the poles.
Incorporating the correct spherical geometry, our SO3ViT model shows a more uniform error distribution.

The following sections
introduce $\mathrm{SO}(3)$-RoPE (Sec.~\ref{sec:SO3RoPE}),
our implementation (Sec.~\ref{sec:implementation}),
and experiments (Sec.~\ref{sec:experiments}).
More related work, mathematical definitions, and theorems are in the Appendix.

\section{SO(3) Rotary Position Embedding}
\label{sec:SO3RoPE}

Our transformers process feature maps $f:\Stwo \rightarrow \mathbb{R}^c$ that lie on the surface of the sphere \Stwo{}.
For instance, our experiments predict the dynamics of the shallow water equation on $S^2$,
whose state is given by spherical velocity fields and fluid heights.
In practice, the input signal often comes discretised on a point cloud or grid,
and is represented by a feature tensor $F\in\mathbb{R}^{N \times c}$
and Euclidean positions $P \in \mathbb{R}^{N\times 3}$ on the unit sphere $S^2 \subset \mathbb{R}^3$.

A key property of the spherical geometry is its rotational symmetry.
Our networks are designed to be \SOthree{} rotation equivariant --
this ensures that they respect the spherical geometry,
treating tokens at any locations equivalently,
independent from their absolute positions.
Intuitively, \SOthree{}-equivariance means that moving a spherical input signal results in an equivalent movement of the predicted output.
Formally, such movements are given by \SOthree{} group actions $[R \rhd f](s) := f(R^{-1}s)$ on spherical signals, where $R\in \SOthree{}$.
An operator or network $\mathcal{T}$ is called \SOthree{} equivariant if it commutes with these actions, that is, if
$\mathcal{T}[R \rhd f] = R \rhd \mathcal{T}[f]$ for any $R\in \mathrm{SO}(3)$.

As in standard ViTs, we tokenise the sphere via convolutional patch embeddings.
Following spherical CNNs, we embed spherical signals via so-called \emph{lifting convolutions} (Def.~\ref{def:lifting_conv}) \citep{Cohen2018SphericalC,pmlr-v80-kondor18a, weiler2023EquivariantAndCoordinateIndependentCNNs}.
Applying a spherical kernel in any \SOthree{}-pose, these convolutions result in tokens on the group \SOthree{}.
In our implementation, we sample a discrete set of $\SOthree{}$ poses for patch tokenisation.
All our tokens are therefore defined on the lifted domain \SOthree{} rather than the input signal domain $S^2$.

To inform the attention operation about spherical geometry we positionally embed our tokens.
Following RoPE, we multiply token features $f(R)$ at $R\in\SOthree{}$ by a position-dependent unitary matrix $\pi(R)\in \mathbb{R}^{c\times c}$.
In standard RoPE one uses position-dependent \mbox{$2\!\times\!2$} rotation matrices,
corresponding to the unitary representations of the 1D translational group. 
Analogously, \SOthree{}-RoPE acts via \emph{unitary \SOthree{} representations} $\pi$.
Due to unitarity and the composition property we get
\begin{equation}
\label{eqn:group_rep}
\pi(U)^\top \pi(V)\ =\ \pi(U)^{-1} \pi(V)\ =\ \pi(U^{-1} V)
\qquad\textup{for any }\quad U,V \in \SOthree{}.
\end{equation}

\begin{bdefinition}[\SOthree{}-RoPE]
\label{def:rope}
Let $f: \SOthree{} \rightarrow  \mathbb{R}^c$ be tokens on $\mathrm{SO}(3)$
and \mbox{$\pi: \SOthree{}\rightarrow \mathbb{R}^{c \times c}$} any unitary \SOthree{}-representation (Def. \ref{def:unitary_so3_rep}).
Our \SOthree{}-RoPE embedding operator $\mathbf{\Upsilon}$ is given by 

\begin{equation}
    \mathbf{\Upsilon}\big[f\big](R)\ :=\ \pi(R)\, f(R) \qquad\textup{for any }\ \ R\in \SOthree{}
\end{equation}
\end{bdefinition}

As in conventional RoPE, we embed our queries and keys, however, in our case on \SOthree{}.
The attention inner product between query $q(U)$ at $U$ and key $k(V)$ at $V$ depends then only on their relative \SOthree{}-position $U^{-1}V$:
\begin{equation}
\label{eqn:grope_rel}
    \mathbf{\Upsilon}\big[q\big](U)^\top\, \mathbf{\Upsilon}\big[k\big](V)
    \ \overset{\text{Def.~\ref{def:rope}}}{=}\ q(U)^\top \pi(U)^\top\, \pi(V)\, k(V)
    \ \overset{(\ref{eqn:group_rep})}{=}\ q(U)^\top\, \pi(U^{-1}V)\, k(V)
\end{equation}

As a result, rotating the queries $R \rhd q$ and keys $R\rhd k$ leads to accordingly rotated attention scores:
\begin{equation}
\begin{aligned}
    \mathbf{\Upsilon}\big[R \rhd q\big](U)^\top\, \mathbf{\Upsilon}\big[R \rhd k\big](V)
    &\ \overset{(\ref{eqn:grope_rel})}{=}\ \big[R \rhd q\big](U)^\top\, \pi(U^{-1}V)\, \big[R \rhd k\big](V)\\
    &\ =\ q(R^{-1}U)^\top\, \pi(U^{-1}V)\, k(R^{-1}V)\\
    &\ =\ q(R^{-1}U)^\top\, \pi(U^{-1}R\, R^{-1} V)\, k(R^{-1}V)\\
    &\ \overset{(\ref{eqn:grope_rel})}{=}\ \mathbf{\Upsilon}\big[q\big](R^{-1}U)^\top\, \mathbf{\Upsilon}\big[k\big](R^{-1}V)
\end{aligned}
\end{equation}

Therefore our \SOthree{}-RoPE embedding produces a rotation-equivariant attention.

\begin{btheorem}[\SOthree{}-RoPE Attention is equivariant] 
\label{thm:rope_equivariant}
   Let $q,k,v: \SOthree{} \rightarrow \mathbb{R}^c$ be query, key and value functions, and $\operatorname{attn}[q,k,v]$ be the dot-product attention operator (Def.~\ref{def:cont_attn}). Then 
   \begin{equation}
      \operatorname{attn}\big[\mathbf{\Upsilon}[R \rhd q],\, \mathbf{\Upsilon}[R \rhd k],\, R \rhd v \big]
      \ =\ R \rhd \operatorname{attn}\big[\mathbf{\Upsilon}[q],\, \mathbf{\Upsilon}[k],\, v\big]
      \qquad\textup{for any }\ \ R\in \SOthree{}.
   \end{equation}
\end{btheorem}

Theorem~\ref{thm:rope_equivariant} is proved in Appendix~\ref{app:proofs}.

\section{Implementation}
\label{sec:implementation}

\paragraph{Patch embedding \& Lifting.}

Our spherical input signals are sampled on an icosphere,
constructed by recursively subdividing the triangular faces of an icosahedron \citep{jiang2018spherical,zhang2019orientation,Cohen2019GaugeEC}.
This produces a hierarchical, nearly uniform discretisation of \Stwo{}.

To embed patches, we convolve triangular kernels on the icospherical grid,
reducing resolution to a few tokens and lifting to \SOthree.
Specifically, we use the grid's hierarchical structure to divide it into $M$ non-overlapping triangle patches
and pick the corresponding \SOthree{}-poses $R_1,\dots,R_{3M}$ such that triangle kernels land exactly on every patch in each of three possible directions, thus resulting in $3M$ tokens.
Assuming our triangle patches are roughly equilateral
allows us to compute this lifting convolution without interpolation.
We store the tokens' \SOthree{} positions $R(\alpha,\beta,\gamma)=R_z(\alpha)R_y(\beta)R_z(\gamma)$ via their Euler angles $\alpha,\beta,\gamma$.

\paragraph{Unpatchification.}
For dense predictions on the original icospherical grid we apply a final transposed convolution.
It unpatchifies tokens via resolution-increasing triangular kernels (Def. \ref{def:proj_conv}).
\paragraph{SO(3) RoPE.}
As unitary representation
$\pi : \SOthree{} \rightarrow \mathbb{R}^{c\times c}$
for token embedding we use a block-diagonal operator built from Wigner D-matrices
$D^{\ell}(R) \in \mathbb{R}^{(2l+1) \times (2l+1)}$ \citep{wigner1931gruppentheorie, weiler2023EquivariantAndCoordinateIndependentCNNs}.
We pick each order $l$ with multiplicity one, choosing as many blocks as our embedding dimension $c$ allows:
\begin{equation}
    \pi\big(R(\alpha,\beta,\gamma)\big) \ =\ 
\begin{pmatrix}
D^0(\alpha,\beta,\gamma) & 0   & \cdots & 0 \\
0   & D^1(\alpha,\beta,\gamma) & \cdots & 0 \\
\vdots & \vdots & \ddots & \vdots \\
0 & 0 & \cdots & D^L(\alpha,\beta,\gamma)
\end{pmatrix}
\end{equation}
Irreducible representations are the smallest possible building blocks,
allowing for a cheap encoding.
After applying \SOthree{}-RoPE, we use the standard FlashAttention   \citep{dao2022flashattention} implementation in PyTorch to compute self-attention \citep{paszke2017automatic}.

\paragraph{Equivariance.}
Only attention layers mix information between tokens; all other layers act token-wise.
With \SOthree{}-RoPE, attention depends only on relative poses and is therefore equivariant to continuous rotations in \SOthree{} (Theorem~\ref{thm:rope_equivariant}).
The lifting layer, however, relies on the icosphere, so the full model is exactly equivariant only to the icosahedral subgroup of \SOthree{}, which is commonly used to discretize rotations \citep{cesa2021ENsteerable,Cohen2019GaugeEC}.
We verify numerically that a randomly initialised model is equivariant to all 60 rotations up to floating-point precision (Appendix~\ref{app:equivariance}).

\paragraph{Symmetry breaking by longitude encoding.}
Not all spherical problems are fully \SOthree{}-equivariant --
instead, the symmetry might be broken to rotations around the $z$-axis  \citep{Amato2025CMBMLAC,armeni2017joint,Bonev2023SphericalFN}.
In such cases, we use an additive absolute position embedding of the longitude, summed to spherical input signals.
Specifically, the longitude angle $\beta$ is encoded via Legendre polynomials $[P^1(\cos\beta),...,P^c(\cos\beta)]^\top$.

\label{others}

\section{Experiments}
\label{sec:experiments}

\subsection{Setup}

\paragraph{Dataset.}
We choose the shallow water equation on a rotating sphere as our benchmark,
using the same on-the-fly data generation and evaluation metrics proposed in \citep{Bonev2025AttentionOT}.
Due to the sphere's rotation there is a Coriolis force.
It breaks the full \SOthree{} symmetry of the sphere down to the subgroup of \SOtwo{} rotations around the $z$-axis. 

\paragraph{Baseline.}
As a baseline model we pick the S2Transformer \citep{Bonev2025AttentionOT} with local neighborhood attention.
DISCO convolutions \citep{Ocampo2022ScalableAE} patchify the sphere to tokens that lie on an equiangular grid.
Positions are embedded via additive spherical harmonics embeddings
--- this is an \emph{absolute} position encoding, thus breaking equivariance.

\paragraph{Training.}
We train for 100 epochs, using the Adam optimiser \citep{kingma2014adam} and a ReduceOnPlateau scheduler, as in \citet{Bonev2025AttentionOT}.
In addition, we test the models also with a cosine scheduler.
We do a grid search for the number of layers [4,8,16,32,64], the number of heads [1,2,4,8] and the embedding dimension [128,256,512].
Per layer and embedding dimension we pick the number of heads and scheduler based on the best $\ell_2$ error.
We train on a single A100 with 40GB memory and a batch size of 16.
Configuration consuming too much memory are skipped.

\subsection{Results}\label{sec: Results}
\vspace*{-4pt}

\begin{figure}[t]
    \centering
    \begin{minipage}[t]{3.3in}
        \centering
        \includegraphics[]{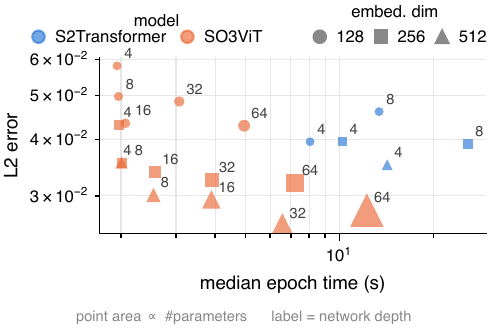}
        \caption{The $\ell_2$ error of predicting the next simulation timestep vs. the median epoch time. SO3ViT outperforms S2Transformer \citep{Bonev2025AttentionOT} given the same computational budget.}
        \label{fig:scaling}
    \end{minipage}
    \hfill
    \begin{minipage}[t]{2.1in}
        \centering
        \includegraphics[width=2.1in]{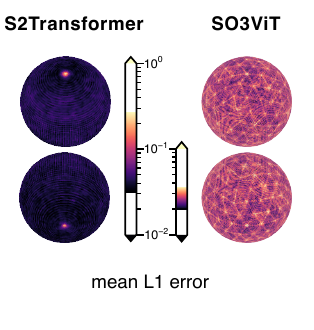}
        \caption{SO3ViT has lower and more homogeneously distributed prediction errors than S2Transformer.}
        \label{fig:l1_error}
    \end{minipage}
\vspace*{-4pt}
\end{figure}

\paragraph{Accuracy-Efficiency Trade-off.}
We train multiple configurations of S2Transformer and SO3ViT.
Figure~\ref{fig:scaling} plots the trained models' $\ell_2$ prediction errors versus their median epoch times.
Our models achieve lower errors at comparable wall-clock times,
and lower wall-clock times at each error level,
forming a significantly improved Pareto frontier.  
This is despite the fact that the S2Transformer uses the torch-harmonics \citep{Bonev2023SphericalFN,Bonev2025AttentionOT} implementations with customised CUDA kernels. We attribute the efficiency of SO3ViT to the use of FlashAttention and the icosphere discretisation. The equiangular discretisation of the S2Transformer oversamples the poles resulting in wasted tokens, whereas our icosphere discretisation samples the sphere uniformly, allowing us to use fewer tokens. SO3ViT also requires less memory and achieves faster inference than S2Transformer. We provide a detailed analysis in Appendix~\ref{app:memory_speed}.

Both models benefit from scaling parameters, although S2Transformer saturates at higher parameter counts.
SO3ViT benefits more from scaling its embedding dimension rather than scaling its depth.
This might be due to larger embedding dimensions allowing for a higher maximal order of Wigner D-matrices, and therefore positional resolution in our \SOthree{}-RoPE embeddings.

\paragraph{Error distribution.}
Next we investigate the spatial structure of prediction errors, visualized in Figure~\ref{fig:l1_error}.
For each point and model, we compute the average error over 200 simulations.
S2Transformer shows an inhomogeneous error distribution, peaking around the poles.
This is caused by two design choices:
First, they operate on an equirectangular grid, which is inhomogeneous with diverging sampling density towards the poles.
Moreover, their patch embedding utilizes a variant of DISCO convolutions which is merely \SOtwo{}-equivariant rather than \SOthree{}-equivariant.
As discussed in \citep{weiler2023EquivariantAndCoordinateIndependentCNNs}, this approach suffers inevitably from topological discontinuities around the poles.
Our SO3ViT, on the other hand, shows a more symmetric error distribution.
One can discern the icosahedral grid pattern, which is due to slight triangle distortions when mapping the icosahedron to an icosphere.
Note that the two plots use a different color range, with the largest error magnitude of SO3ViT corresponding roughly to the smallest error magnitude of S2Transformer.
These results underline the importance of building an architecture that truly respects the geometry of the data domain.

\vspace*{-2pt}
\section{Conclusion}
\vspace*{-6pt}
We proposed \SOthree{}-RoPE, a relative position encoding for spherical transformers, acting via unitary \SOthree{}-irreps.
It is FlashAttention-compatible and outperforms S2Transformer at equal compute.

\subsubsection*{Acknowledgements.}
This research is funded by the Deutsche Forschungsgemeinschaft (DFG, German Research
Foundation), project number 501927736, within the DFG Priority Programme 2353: Daring More Intelligence - Design Assistants in Mechanics and Dynamics’.
This project has received funding from the European Research Council (ERC) under the European Union’s Horizon Europe research and innovation programme (Grant agreement No. 101041669).
The authors gratefully acknowledge the computing time granted by the Resource Allocation Board and provided on the supercomputer Emmy/Grete at NHR-Nord@Göttingen as part of the NHR infrastructure. The calculations for this research were conducted with computing resources under the project nii00194.

\bibliographystyle{plainnat}
\bibliography{bib}

@inproceedings{Shen2022PanoFormerPT,
  title={PanoFormer: Panorama Transformer for Indoor 360° Depth Estimation},
  author={Zhijie Shen and Chunyu Lin and Kang Liao and Lang Nie and Zishuo Zheng and Yao Zhao},
  booktitle={European Conference on Computer Vision},
  year={2022}
}

@article{Kohler2025BridgingEG,
  title={Bridging Equivariant GNNs and Spherical CNNs for Structured Physical Domains},
  author={Colin Kohler and Purvik Patel and Nathan Vaska and Justin Goodwin and Matthew C. Jones and Robert Platt and Rajmonda Caceres and Robin Walters},
  journal={Advances in Neural Information Processing Systems 38},
  year={2025}
}

@article{Defferrard2020DeepSphereAG,
  title={DeepSphere: a graph-based spherical CNN},
  author={Micha{\"e}l Defferrard and Martino Milani and Fr{\'e}d{\'e}rick Gusset and Nathanael Perraudin},
  journal={ArXiv},
  year={2020},
  volume={abs/2012.15000}
}

@article{Benny2024SphereUFormerAU,
  title={SphereUFormer: A U-Shaped Transformer for Spherical 360 Perception},
  author={Yaniv Benny and Lior Wolf},
  journal={2025 IEEE/CVF Conference on Computer Vision and Pattern Recognition (CVPR)},
  year={2024},
  pages={940-950}
}

@article{Perraudin2018DeepSphereES,
  title={DeepSphere: Efficient spherical Convolutional Neural Network with HEALPix sampling for cosmological applications},
  author={Nathanael Perraudin and Micha{\"e}l Defferrard and Tomasz Kacprzak and Raphael Sgier},
  journal={Astron. Comput.},
  year={2018},
  volume={27},
  pages={130-146}
}

@article{Zhdanov2026SparseAT,
  title={(Sparse) Attention to the Details: Preserving Spectral Fidelity in ML-based Weather Forecasting Models},
  author={Maksim Zhdanov and Ana Lucic and Max Welling and Jan-Willem van de Meent},
  journal={ArXiv},
  year={2026},
  volume={abs/2604.16429}
}

@article{Thomas2018TensorFN,
  title={Tensor Field Networks: Rotation- and Translation-Equivariant Neural Networks for 3D Point Clouds},
  author={Nathaniel Thomas and Tess E. Smidt and Steven M. Kearnes and Lusann Yang and Li Li and Kai Kohlhoff and Patrick F. Riley},
  journal={ArXiv},
  year={2018},
  volume={abs/1802.08219}
}

@article{Cohen2018SphericalC,
  title={Spherical CNNs},
  author={Taco Cohen and Mario Geiger and Jonas K{\"o}hler and Max Welling},
  journal={ArXiv},
  year={2018},
  volume={abs/1801.10130}
}

@article{Bonev2025AttentionOT,
  title={Attention on the Sphere},
  author={Boris Bonev and Max Rietmann and Andr{\'e}anne Paris and Alberto Carpentieri and Thorsten Kurth},
  journal={ArXiv},
  year={2025},
  volume={abs/2505.11157}
}

@article{Islam2025PlatonicTA,
  title={Platonic Transformers: A Solid Choice For Equivariance},
  author={Mohammad Mohaiminul Islam and Rishabh Anand and David R. Wessels and Friso de Kruiff and Thijs P. Kuipers and Rex Ying and Clara I. S'anchez and Sharvaree P. Vadgama and Georg B{\"o}kman and Erik J. Bekkers},
  journal={ArXiv},
  year={2025},
  volume={abs/2510.03511}
}

@article{Zhang2024SGFormerSG,
  title={SGFormer: Spherical Geometry Transformer for 360 Depth Estimation},
  author={Junsong Zhang and Zisong Chen and Chunyu Lin and Lang Nie and Zhijie Shen and Junda Huang and Yao Zhao},
  journal={ArXiv},
  year={2024},
  volume={abs/2404.14979}
}

@inproceedings{Bonev2023SphericalFN,
  title={Spherical Fourier Neural Operators: Learning Stable Dynamics on the Sphere},
  author={Boris Bonev and Thorsten Kurth and Christian Hundt and Jaideep Pathak and Maximilian Baust and Karthik Kashinath and Anima Anandkumar},
  booktitle={International Conference on Machine Learning},
  year={2023}
}

@article{Tang2026NeuralOF,
  title={Neural Operators for Biomedical Spherical Heterogeneity},
  author={Hao Tang and Hao Chen and Hao Li and Chao Li},
  journal={ArXiv},
  year={2026},
  volume={abs/2601.03561}
}

@article{Carlsson2023HEALSWINAV,
  title={HEAL-SWIN: A Vision Transformer on the Sphere},
  author={Oscar Carlsson and Jan E. Gerken and Hampus Linander and Heiner Spiess and Fredrik Ohlsson and Christoffer Petersson and Daniel Persson},
  journal={2024 IEEE/CVF Conference on Computer Vision and Pattern Recognition (CVPR)},
  year={2023},
  pages={6067-6077}
}

@article{Ocampo2022ScalableAE,
  title={Scalable and Equivariant Spherical CNNs by Discrete-Continuous (DISCO) Convolutions},
  author={Jeremy Ocampo and Matthew Alexander Price and Jason D. McEwen},
  journal={ArXiv},
  year={2022},
  volume={abs/2209.13603}
}

@article{Cohen2019GaugeEC,
  title={Gauge Equivariant Convolutional Networks and the Icosahedral CNN},
  author={Taco Cohen and Maurice Weiler and Berkay Kicanaoglu and Max Welling},
  journal={ArXiv},
  year={2019},
  volume={abs/1902.04615}
}

@article{Amato2025CMBMLAC,
  title={CMB-ML: A Cosmic Microwave Background Dataset for the Oldest Possible Computer Vision Task},
  author={James Amato and Yu Xie and Leonel Medina-Varela and Ammar Aljerwi and Adam McCutcheon and T. Seth Rippentrop and Kristian Gonzalez and Jacques Delabrouille and Mustapha Ishak and Nicholas Ruozzi},
  journal={2025 IEEE/CVF International Conference on Computer Vision (ICCV)},
  year={2025},
  pages={9418-9430}
}

@inproceedings{cohen2016group,
  title={Group equivariant convolutional networks},
  author={Cohen, Taco and Welling, Max},
  booktitle={International conference on machine learning},
  pages={2990--2999},
  year={2016},
  organization={PMLR}
}

@article{vaswani2017attention,
  title={Attention is all you need},
  author={Vaswani, Ashish and Shazeer, Noam and Parmar, Niki and Uszkoreit, Jakob and Jones, Llion and Gomez, Aidan N and Kaiser, {\L}ukasz and Polosukhin, Illia},
  journal={Advances in neural information processing systems},
  volume={30},
  year={2017}
}

@article{dosovitskiy2020image,
  title={An image is worth 16x16 words: Transformers for image recognition at scale},
  author={Dosovitskiy, Alexey and Beyer, Lucas and Kolesnikov, Alexander and Weissenborn, Dirk and Zhai, Xiaohua and Unterthiner, Thomas and Dehghani, Mostafa and Minderer, Matthias and Heigold, Georg and Gelly, Sylvain and others},
  journal={arXiv preprint arXiv:2010.11929},
  year={2020}
}

@article{bronstein2017geometric,
  title={Geometric deep learning: going beyond euclidean data},
  author={Bronstein, Michael M and Bruna, Joan and LeCun, Yann and Szlam, Arthur and Vandergheynst, Pierre},
  journal={IEEE Signal Processing Magazine},
  volume={34},
  number={4},
  pages={18--42},
  year={2017},
  publisher={IEEE}
}

@book{weiler2026equivariant,
  title={Equivariant and Coordinate Independent Convolutional Networks: A Gauge Field Theory of Neural Networks},
  author={Weiler, Maurice and Forr{\'e}, Patrick and Verlinde, Erik and Welling, Max},
  year={2026},
  publisher={World Scientific}
}

@inproceedings{radford2023robust,
  title={Robust speech recognition via large-scale weak supervision},
  author={Radford, Alec and Kim, Jong Wook and Xu, Tao and Brockman, Greg and McLeavey, Christine and Sutskever, Ilya},
  booktitle={International conference on machine learning},
  pages={28492--28518},
  year={2023},
  organization={PMLR}
}

@article{su2024roformer,
  title={Roformer: Enhanced transformer with rotary position embedding},
  author={Su, Jianlin and Ahmed, Murtadha and Lu, Yu and Pan, Shengfeng and Bo, Wen and Liu, Yunfeng},
  journal={Neurocomputing},
  volume={568},
  pages={127063},
  year={2024},
  publisher={Elsevier}
}

@article{jumper2021highly,
  title={Highly accurate protein structure prediction with AlphaFold},
  author={Jumper, John and Evans, Richard and Pritzel, Alexander and Green, Tim and Figurnov, Michael and Ronneberger, Olaf and Tunyasuvunakool, Kathryn and Bates, Russ and {\v{Z}}{\'\i}dek, Augustin and Potapenko, Anna and others},
  journal={nature},
  volume={596},
  number={7873},
  pages={583--589},
  year={2021},
  publisher={Nature Publishing Group UK London}
}

@inproceedings{miyato2024gta,
  title={Gta: A geometry-aware attention mechanism for multi-view transformers},
  author={Miyato, Takeru and Jaeger, Bernhard and Welling, Max and Geiger, Andreas},
  booktitle={International Conference on Learning Representations},
  volume={2024},
  pages={8172--8208},
  year={2024}
}

@article{cohen2016steerable,
  title={Steerable cnns},
  author={Cohen, Taco S and Welling, Max},
  journal={arXiv preprint arXiv:1612.08498},
  year={2016}
}

@article{weiler2019general,
  title={General e (2)-equivariant steerable cnns},
  author={Weiler, Maurice and Cesa, Gabriele},
  journal={Advances in neural information processing systems},
  volume={32},
  year={2019}
}

@inproceedings{cesa2021ENsteerable,
    title     = {{A program to build E(N)-equivariant steerable CNNs}},
    author    = {Cesa, Gabriele and Lang, Leon and Weiler, Maurice},
    booktitle = {{International Conference on Learning Representations (ICLR)}},
    url       = {https://openreview.net/forum?id=WE4qe9xlnQw},
    year      = {2022}
}

@inproceedings{shaw2018self,
  title={Self-attention with relative position representations},
  author={Shaw, Peter and Uszkoreit, Jakob and Vaswani, Ashish},
  booktitle={Proceedings of the 2018 Conference of the North American Chapter of the Association for Computational Linguistics: Human Language Technologies, Volume 2 (Short Papers)},
  pages={464--468},
  year={2018}
}

@article{armeni2017joint,
  title={Joint 2d-3d-semantic data for indoor scene understanding},
  author={Armeni, Iro and Sax, Sasha and Zamir, Amir R and Savarese, Silvio},
  journal={arXiv preprint arXiv:1702.01105},
  year={2017}
}

@article{rasp2024weatherbench,
  title={WeatherBench 2: A benchmark for the next generation of data-driven global weather models},
  author={Rasp, Stephan and Hoyer, Stephan and Merose, Alexander and Langmore, Ian and Battaglia, Peter and Russell, Tyler and Sanchez-Gonzalez, Alvaro and Yang, Vivian and Carver, Rob and Agrawal, Shreya and others},
  journal={Journal of Advances in Modeling Earth Systems},
  volume={16},
  number={6},
  pages={e2023MS004019},
  year={2024},
  publisher={Wiley Online Library}
}

@inproceedings{kazemnejad2023impact,
  title={The Impact of Positional Encoding on Length Generalization in Transformers},
  author={Kazemnejad, Amirhossein and Padhi, Inkit and Natesan Ramamurthy, Karthikeyan and Das, Payel and Reddy, Siva},
  booktitle={Advances in Neural Information Processing Systems (NeurIPS)},
  volume={36},
  year={2023}
}

@article{kogkalidis2024algebraic,
  title={Algebraic positional encodings},
  author={Kogkalidis, Konstantinos and Bernardy, Jean-Philippe and Garg, Vikas},
  journal={Advances in Neural Information Processing Systems},
  volume={37},
  pages={34824--34845},
  year={2024}
}

@article{dao2022flashattention,
  title={Flashattention: Fast and memory-efficient exact attention with io-awareness},
  author={Dao, Tri and Fu, Dan and Ermon, Stefano and Rudra, Atri and R{\'e}, Christopher},
  journal={Advances in neural information processing systems},
  volume={35},
  pages={16344--16359},
  year={2022}
}

@inproceedings{paszke2017automatic,
  title={Automatic differentiation in PyTorch},
  author={Paszke, Adam and Gross, Sam and Chintala, Soumith and Chanan, Gregory and Yang, Edward and DeVito, Zachary and Lin, Zeming and Desmaison, Alban and Antiga, Luca and Lerer, Adam},
  booktitle={NIPS-W},
  year={2017}
}

@InProceedings{pmlr-v80-kondor18a,
  title = 	 {On the Generalization of Equivariance and Convolution in Neural Networks to the Action of Compact Groups},
  author =       {Kondor, Risi and Trivedi, Shubhendu},
  booktitle = 	 {Proceedings of the 35th International Conference on Machine Learning},
  pages = 	 {2747--2755},
  year = 	 {2018},
  editor = 	 {Dy, Jennifer and Krause, Andreas},
  volume = 	 {80},
  series = 	 {Proceedings of Machine Learning Research},
  month = 	 {10--15 Jul},
  publisher =    {PMLR},
  url = 	 {https://proceedings.mlr.press/v80/kondor18a.html}
}

@book{weiler2023EquivariantAndCoordinateIndependentCNNs,
    title     = {{Equivariant and Coordinate Independent Convolutional Networks}},
    subtitle  = {{A Gauge Field Theory of Neural Networks}},
    author    = {Weiler, Maurice and Forré, Patrick and Verlinde, Erik and Welling, Max},
    url       = {https://maurice-weiler.gitlab.io/cnn_book/EquivariantAndCoordinateIndependentCNNs.pdf},
    year      = {2023}
}

@article{wigner1931gruppentheorie,
  title={Gruppentheorie und ihre Anwendung auf die Quantenmechanik der Atomspektren},
  author={Wigner, Eugen},
  year={1931},
  publisher={Springer}
}

@article{kingma2014adam,
  title={Adam: A method for stochastic optimization},
  author={Kingma, Diederik P and Ba, Jimmy},
  journal={arXiv preprint arXiv:1412.6980},
  year={2014}
}

@article{cohen2019general,
  title={A general theory of equivariant cnns on homogeneous spaces},
  author={Cohen, Taco S and Geiger, Mario and Weiler, Maurice},
  journal={Advances in neural information processing systems},
  volume={32},
  year={2019}
}

@inproceedings{
jiang2018spherical,
title={Spherical {CNN}s on Unstructured Grids},
author={Chiyu Max Jiang and Jingwei Huang and Karthik Kashinath and Prabhat and Philip Marcus and
        Matthias Niessner},
booktitle={International Conference on Learning Representations},
year={2019},
url={https://openreview.net/forum?id=Bkl-43C9FQ},
}

@inproceedings{zhang2019orientation,
  title={Orientation-aware semantic segmentation on icosahedron spheres},
  author={Zhang, Chao and Liwicki, Stephan and Smith, William and Cipolla, Roberto},
  booktitle={2019 IEEE/CVF International Conference on Computer Vision (ICCV)},
  pages={3532--3540},
  year={2019},
  organization={IEEE}
}

@inproceedings{esteves2018learning,
  title={Learning so (3) equivariant representations with spherical cnns},
  author={Esteves, Carlos and Allen-Blanchette, Christine and Makadia, Ameesh and Daniilidis, Kostas},
  booktitle={Proceedings of the european conference on computer vision (ECCV)},
  pages={52--68},
  year={2018}
}

@inproceedings{kondor2018generalization,
  title={On the generalization of equivariance and convolution in neural networks to the action of compact groups},
  author={Kondor, Risi and Trivedi, Shubhendu},
  booktitle={International conference on machine learning},
  pages={2747--2755},
  year={2018},
  organization={PMLR}
}

@article{bronstein2021geometric,
  title={Geometric deep learning: Grids, groups, graphs, geodesics, and gauges},
  author={Bronstein, Michael M and Bruna, Joan and Cohen, Taco and Veli{\v{c}}kovi{\'c}, Petar},
  journal={arXiv preprint arXiv:2104.13478},
  year={2021}
}

@article{driscoll1994computing,
  title={Computing Fourier transforms and convolutions on the 2-sphere},
  author={Driscoll, James R and Healy, Dennis M},
  journal={Advances in applied mathematics},
  volume={15},
  number={2},
  pages={202--250},
  year={1994},
  publisher={Elsevier}
}

@article{weiler20183d,
  title={3d steerable cnns: Learning rotationally equivariant features in volumetric data},
  author={Weiler, Maurice and Geiger, Mario and Welling, Max and Boomsma, Wouter and Cohen, Taco},
  journal={Advances in Neural information processing systems},
  volume={31},
  year={2018}
}

@article{kondor2018clebsch,
  title={Clebsch--gordan nets: a fully fourier space spherical convolutional neural network},
  author={Kondor, Risi and Lin, Zhen and Trivedi, Shubhendu},
  journal={Advances in neural information processing systems},
  volume={31},
  year={2018}
}

@article{kostelec2008ffts,
  title={FFTs on the rotation group},
  author={Kostelec, Peter J and Rockmore, Daniel N},
  journal={Journal of Fourier analysis and applications},
  volume={14},
  number={2},
  pages={145--179},
  year={2008},
  publisher={Springer}
}

@article{gorski2005healpix,
  title={HEALPix: A framework for high-resolution discretization and fast analysis of data distributed on the sphere},
  author={Gorski, Krzysztof M and Hivon, Eric and Banday, Anthony J and Wandelt, Benjamin D and Hansen, Frode K and Reinecke, Mstvos and Bartelmann, Matthia},
  journal={The Astrophysical Journal},
  volume={622},
  number={2},
  pages={759--771},
  year={2005}
}

@inproceedings{ling2023panoswin,
  title={PanoSwin: a pano-style swin transformer for panorama understanding},
  author={Ling, Zhixin and Xing, Zhen and Zhou, Xiangdong and Cao, Manliang and Zhou, Guichun},
  booktitle={2023 IEEE/CVF Conference on Computer Vision and Pattern Recognition (CVPR)},
  pages={17755--17764},
  year={2023},
  organization={IEEE}
}

@inproceedings{yun2023egformer,
  title={Egformer: Equirectangular geometry-biased transformer for 360 depth estimation},
  author={Yun, Ilwi and Shin, Chanyong and Lee, Hyunku and Lee, Hyuk-Jae and Rhee, Chae Eun},
  booktitle={2023 IEEE/CVF International Conference on Computer Vision (ICCV)},
  pages={6078--6089},
  year={2023},
  organization={IEEE}
}

@article{van2026circular,
  title={A Circular Argument: Does RoPE need to be Equivariant for Vision?},
  author={van de Geijn, Chase and L{\"u}ddecke, Timo and Turishcheva, Polina and Ecker, Alexander},
  journal={Advances in Neural Information Processing Systems},
  volume={38},
  pages={20388--20424},
  year={2026}
}

@article{liu2025rethinking,
  title={Rethinking rope: A mathematical blueprint for n-dimensional positional embedding},
  author={Liu, Haiping and Lin, Lijing and Sun, Jingyuan and Shangguan, Zhegong and Alvarez, Mauricio A and Zhou, Hongpeng},
  journal={arXiv preprint arXiv:2504.06308},
  year={2025}
}

@inproceedings{coors2018spherenet,
  title={Spherenet: Learning spherical representations for detection and classification in omnidirectional images},
  author={Coors, Benjamin and Condurache, Alexandru Paul and Geiger, Andreas},
  booktitle={European conference on computer vision},
  pages={525--541},
  year={2018},
  organization={Springer}
}

@article{su2017learning,
  title={Learning spherical convolution for fast features from 360 imagery},
  author={Su, Yu-Chuan and Grauman, Kristen},
  journal={Advances in neural information processing systems},
  volume={30},
  year={2017}
}

@article{boomsma2017spherical,
  title={Spherical convolutions and their application in molecular modelling},
  author={Boomsma, Wouter and Frellsen, Jes},
  journal={Advances in neural information processing systems},
  volume={30},
  year={2017}
}

@inproceedings{wu2018squeezeseg,
  title={Squeezeseg: Convolutional neural nets with recurrent crf for real-time road-object segmentation from 3d lidar point cloud},
  author={Wu, Bichen and Wan, Alvin and Yue, Xiangyu and Keutzer, Kurt},
  booktitle={2018 IEEE international conference on robotics and automation (ICRA)},
  pages={1887--1893},
  year={2018},
  organization={IEEE}
}

@inproceedings{yogamani2019woodscape,
  title={Woodscape: A multi-task, multi-camera fisheye dataset for autonomous driving},
  author={Yogamani, Senthil and Hughes, Ciar{\'a}n and Horgan, Jonathan and Sistu, Ganesh and Varley, Padraig and O'Dea, Derek and Uric{\'a}r, Michal and Milz, Stefan and Simon, Martin and Amende, Karl and others},
  booktitle={Proceedings of the IEEE/CVF international conference on computer vision},
  pages={9308--9318},
  year={2019}
}

@inproceedings{lee2019spherephd,
  title={SpherePHD: Applying CNNs on a spherical PolyHeDron representation of 360° images},
  author={Lee, Yeonkun and Jeong, Jaeseok and Yun, Jongseob and Cho, Wonjune and Yoon, Kuk-Jin},
  booktitle={2019 IEEE/CVF Conference on Computer Vision and Pattern Recognition (CVPR)},
  pages={9173--9181},
  year={2019},
  organization={IEEE}
}

@inproceedings{yan2022spheredepth,
  title={Spheredepth: Panorama depth estimation from spherical domain},
  author={Yan, Qingsong and Wang, Qiang and Zhao, Kaiyong and Li, Bo and Chu, Xiaoweo and Deng, Fei},
  booktitle={2022 International Conference on 3D Vision (3DV)},
  pages={1--10},
  year={2022},
  organization={IEEE}
}

\newpage
\appendix
{\LARGE Appendix}
\vspace*{14pt}

\section{Related Work / Background} \label{app:Related Work}

Signals that naturally live on the sphere arise throughout the sciences and engineering, from climate and weather fields defined over the globe \citep{rasp2024weatherbench} and the cosmic microwave background in cosmology \citep{gorski2005healpix}, to omnidirectional and panoramic imagery \citep{su2017learning,coors2018spherenet}, molecular surfaces \citep{boomsma2017spherical}, and data from fisheye or LiDAR sensors \citep{yogamani2019woodscape, wu2018squeezeseg}. Processing such data faithfully requires the spherical geometry to be encoded into the model itself, rather than treated as an afterthought. In this section, we review three strands of work: the established theory of spherical CNNs, recent efforts to bring transformers to the sphere, and general advances in encoding geometry into transformers through positional encodings.
 
\paragraph{Spherical convolutions.}
Geometric Deep Learning (GDL)\citep{bronstein2017geometric, weiler2026equivariant} studies how to incorporate data geometry into deep learning models. The guiding principle of GDL is that of the \textit{Erlangen programme}\citep{bronstein2021geometric} which characterizes geometries through group theory and symmetries. Thus, to faithfully represent the geometry, a model should respect the symmetries of its data. 

For a domain which is a sequence or line, such as in language, the most relevant symmetry is translation.
Convolutions are the only linear operators that are translation-equivariant\citep{kondor2018generalization}.
On the sphere, $S^2$, the relevant symmetries are 3D rotations; any rotation -- or more formally any element of $SO(3)$, i.e. the space of $3\times3$ special (unit determinant) orthogonal matrices -- leaves $S^2$ invariant. Thus, the translations of the common convolution are replaced by rotations. The operator we seek is a linear map that is rotation-equivariant, which turns out to be a spherical convolution \citep{Cohen2018SphericalC, driscoll1994computing}. A well-known obstacle is that a linear operator mapping a signal on $S^2$ back to a signal on $S^2$ can only be $SO(3)$-equivariant when its kernel is isotropic, i.e.\ depends solely on the geodesic distance between points. Such zonal kernels are blind to orientation and therefore limit a model's expressivity.

Spherical CNNs overcome this by lifting features from the sphere to the rotation group\citep{Cohen2018SphericalC, esteves2018learning, Cohen2019GaugeEC}: the first layer maps a signal on $S^2$ to a signal on $SO(3)$, after which subsequent layers convolve on $SO(3)$ itself, where anisotropic and far more expressive kernels remain equivariant. Two complementary realisations of this idea are common. Group convolutions\citep{cohen2016group} enforce equivariance directly in the spatial domain by correlating over transformed copies of the kernel. Steerable convolutions\citep{cohen2016steerable, weiler20183d, weiler2019general,cohen2019general} instead operate in the spectral domain and are the group-theoretic counterpart of computing a convolution through the (generalized) Fourier transform, representing signals through the irreducible representations of the symmetry group. For $SO(3)$ these irreps are the Wigner-$D$ matrices, with the corresponding transform expanding signals in spherical harmonics~\citep{kondor2018clebsch, kostelec2008ffts}.

We leverage both the early works' insights on lifting and the latter works' insights on irreps of $SO(3)$ to construct our positional encoding.

\paragraph{Spherical Transformers.} 
Just as transformers have largely displaced convolutions in vision, a growing line of work adapts them to spherical data.
Before extending a transformer to the sphere, one must first decide how to discretize and represent it.
While planar images allow for regular pixel grids at arbitrary resolutions,
the only regular grids on the sphere correspond to Platonic solids and are of rather coarse resolution.
To achieve higher resolutions, one needs to fall back to irregular sampling schemes:
equirectangular (latitude--longitude) grids, icospheres obtained by repeatedly subdividing an icosahedron, or the equal-area HEALPix pixelisation~\citep{gorski2005healpix} are all common choices. Each carries a different trade-off between sampling uniformity, hierarchical structure, and ease of implementation. In our work, we choose to operate on icosphere subdivisions.

For each sampling scheme, several transformer architectures have been proposed. On the equirectangular grid, the main challenge is counteracting the projection's polar distortion within the attention mechanism itself: PanoFormer~\citep{Shen2022PanoFormerPT} attends over tangent patches with a learnable token flow, EGformer~\citep{yun2023egformer} biases local attention along the panorama's vertical and horizontal directions, and PanoSwin~\citep{ling2023panoswin} adapts the shifted-window design of the Swin transformer to the wrap-around topology. These methods keep the convenience of a rectangular tensor, but must spend architectural effort to undo the distortion it introduces.

Icosphere-based transformers avoid this distortion at the source by operating on a near-uniform mesh. SphereUFormer~\citep{Benny2024SphereUFormerAU} builds a U-shaped hierarchy of local self-attention modules directly on the icosphere, using the recursive subdivision of the mesh to define the pooling and unpooling operations between resolutions. This continues a line of icosahedral networks~\citep{jiang2018spherical, Cohen2019GaugeEC,  lee2019spherephd, zhang2019orientation, Defferrard2020DeepSphereAG, yan2022spheredepth} that exploit the icosahedron's high degree of symmetry, which we likewise build upon.

Others adopt the HEALPix grid, an equal-area, hierarchically nested pixelization, which several transformer-based methods exploit~\citep{Carlsson2023HEALSWINAV, Zhdanov2026SparseAT}. We do not follow this route: while HEALPix offers equal-area sampling and a convenient hierarchical layout, its pixelization lacks the symmetry of the icosphere, which we favor in this work to simplify lifting.

\paragraph{Local attention} A common thread across these architectures is a reliance on local attention, which restricts each query to a neighborhood of nearby tokens rather than the full sphere---both to control the quadratic cost of attention and to supply an inductive bias for locality. On the equirectangular grid this appears as deformable pixel neighborhoods in PanoFormer~\citep{Shen2022PanoFormerPT} and geometry-biased windows in EGformer~\citep{yun2023egformer}, while PanoSwin~\citep{ling2023panoswin}, HEAL-SWIN~\citep{Carlsson2023HEALSWINAV}. SphereUFormer~\citep{Benny2024SphereUFormerAU} likewise attends within mesh neighborhoods on the icosphere. 

By contrast, we control the cost by patchifying analogous to the vanilla ViT. This patchification is one of the reasons we must lift to $SO(3)$. The benefit is that the model captures local dependencies through the patches while the attention focuses on long range.

\paragraph{Positional encodings.}
In a transformer, self-attention is permutation-equivariant and carries no intrinsic notion of position; geometry can therefore enter only through the positional encoding~\citep{kogkalidis2024algebraic} or through geometry-dependent masking such as local or causal attention~\citep{kazemnejad2023impact}. Most modern transformers rely on some form of relative positional encoding (RPE)~\citep{shaw2018self}, which in language corresponds to encoding a translational symmetry~\citep{miyato2024gta, van2026circular, Islam2025PlatonicTA}. The dominant instance is the Rotary Positional Encoding (RoPE)~\citep{su2024roformer}, which has been shown to arise axiomatically from a small set of desirable assumptions on sequential data~\citep{kogkalidis2024algebraic, liu2025rethinking}.

A practical virtue of RoPE is that it is applied to the query and key vectors independently, without materializing a position-dependent attention matrix; this separability is what makes it cheap to compute and easy to extend. Extensions beyond one dimension have, however, focused almost exclusively on toric domains, yielding the widely used axial RoPE for $N$-dimensional inputs~\citep{kogkalidis2024algebraic, liu2025rethinking, van2026circular}. The irreps perspective that explains RoPE's success has not been carried beyond this setting. We take up this perspective and use it to extend RoPE to the sphere---to our knowledge the first positional encoding to do so.

This gap is also visible in existing spherical transformers. Because they typically resort to local attention for tractability~\citep{Benny2024SphereUFormerAU}, they must pair it with a local positional encoding, but these either fall back on an isotropic kernel---echoing the limited expressivity of the zonal kernels of spherical convolutions---or ignore the sphere's gauge problem and thereby cause problems at the poles~\citep{Cohen2019GaugeEC}. Local attention is moreover hardware-inefficient and restricted in expressivity, making such models both slower and less accurate than the patchification scheme we propose.

\section{Memory \& Inference speed}
\label{app:memory_speed}

\begin{table}
\caption{Peak GPU memory during training and inference time. For every S2Transformer configuration, some SO3ViT configuration performs better in $\ell_1$ error, $\ell_2$ error, memory, and inference speed.}
\label{tab:scaling_benchmark}
\begin{tabular}{lrrrrrrr}
\toprule
Model & Depth & Width & Params & $\ell_1$ & $\ell_2$ & Memory & Inference Time \\
\midrule
S2Transformer & 4 & 128 & 0.53~M & 0.031 & 0.040 & 8.01~GB & 141.4~ms \\
S2Transformer & 4 & 256 & 2.12~M & 0.031 & 0.040 & 15.96~GB & 198.8~ms \\
S2Transformer & 4 & 512 & 8.43~M & 0.027 & 0.035 & 31.92~GB & 282.1~ms \\
S2Transformer & 8 & 256 & 4.22~M & 0.031 & 0.039 & 21.99~GB & 471.0~ms \\
\hline
SO3ViT & 4 & 128 & 0.98~M & 0.045 & 0.058 & 0.74~GB & 6.1~ms \\
SO3ViT & 4 & 256 & 3.54~M & 0.034 & 0.043 & 1.28~GB & 9.3~ms \\
SO3ViT & 4 & 512 & 13.37~M & 0.028 & 0.036 & 2.40~GB & 16.2~ms \\
SO3ViT & 16 & 512 & 51.12~M & 0.023 & 0.029 & 8.82~GB & 55.2~ms \\
SO3ViT & 32 & 512 & 101.45~M & \textbf{0.020} & \textbf{0.026} & 17.37~GB & 109.9~ms \\
\bottomrule
\end{tabular}
\end{table}

We compare memory consumption and inference speed of the S2Transformer and SO3ViT (Tab.~\ref{tab:scaling_benchmark}).
At matched accuracy, SO3ViT uses less memory and runs faster, despite having more parameters.

SO3ViT tokenises the input on an icosphere with two subdivisions (320 faces).
The lifting layer then produces $3 \cdot 320 = 960$ tokens.
The S2Transformer instead discretises the input on a $64 \times 128$ equirectangular grid, which gives $8{,}192$ tokens.
Coarser patchification was not possible for the S2Transformer.
Otherwise, it exceeded the 40\,GB memory limit at a batch size of 16.
The reason is the torch-harmonics implementation.
It first upsamples and then applies DISCO convolutions at full output resolution.
These convolutions store an interpolation matrix for every latitude, and its size grows with the kernel size.
Our patchification and unpatchification avoid this cost, since they require no interpolation.
This is possible because our attention mechanism allows a free choice of discretisation of $\SOthree{}$ (Sec.~\ref{sec:implementation}) and our assumption of equilateral triangles of the icosphere.

\section{Equivariance Test}
\label{app:equivariance}

Our \SOthree{}-RoPE attention mechanism is equivariant to continuous rotations in \SOthree{}.
We discretize our data on an icospherical grid, which breaks continuous symmetries to the subgroup of icosahedral symmetries.
This preserves exact icosahedral group equivariance and approximate continuous \SOthree{}-equivariance of the full model.

To verify this, we sample a random input for a randomly initialised model and permute it according to each rotation of the icosahedral group.
We then compare the output for the rotated input with the rotated output for the original input.
For all 60 rotations, the two agree up to floating-point precision.

\section{Definitions}

In order to avoid discretization artifacts and preserve symmetries, all of our definitions are formulated in the continuous setting.
We use scaled dot product attention \cite{vaswani2017attention},
replacing discrete sums over tokens by continuous integrals over \SOthree{}.

\begin{bdefinition}[Continuous Attention]
\label{def:cont_attn}
    Let $q,k,v: \SOthree{} \rightarrow \mathbb{R}^c$.
    We define scaled dot-product attention on \SOthree{} as:
   \begin{equation}
    \operatorname{attn}[q,k,v](U)
    \ =\ \int_{\SOthree{}}\mkern-24mu \operatorname{softmax}\!\big[A[q,k]\big](U,V)\; v(V) \;\mathrm{d}V
    \qquad\textup{for any }\ \ U \in \SOthree{},
    \end{equation}
    with $\operatorname{softmax}[A](U,V) = \frac{e^{A(U,V)}}{\int_{\SOthree{}} e^{A(U,Q)} \,\mathrm{d}Q}$ and the attention score between the query $q(U)$ at $U$ and key $k(V)$ at $V$ given by
    \begin{equation}
        A[q,k](U,V) = q(U)^\top\, k(V)
    \end{equation}
\end{bdefinition}
Note that we absorb the $c$-dependent scaling factor $\frac{1}{\sqrt{c}}$ into the queries and keys; in the implementation we scale the attention scores by it as usual.

Our patch embedding from spherical signals to tokens on $\SOthree{}$ is done via usual lifting group convolutions.
\begin{bdefinition}[Lifting convolution $\Stwo{} \to \SOthree{}$]
\label{def:lifting_conv}
\label{def:lifting-conv}
Let $f_{\text{in}} : \Stwo{} \rightarrow \mathbb{R}^{c_{\text{in}}}$ be a spherical signal and
$\psi_\uparrow : \Stwo{} \rightarrow \mathbb{R}^{c \times c_{\text{in}}}$ an arbitrary filter.
The \emph{lifting convolution} of $f_{\text{in}}$ with $\psi_\uparrow$ is given by:
\begin{equation}
  (\psi_\uparrow \star f_{\text{in}})(R) \ =\ \int_{\Stwo{}} \psi_\uparrow\big(R^{-1}s\big)\, f_{\text{in}}(s) \,\mathrm{d}s
  \qquad\textup{for any }\ \ R \in \SOthree{},
  \label{eq:lifting-conv}
\end{equation}
\end{bdefinition}
Intuitively, this operation applies a spherical kernel in every $\SOthree{}$-rotated pose.

\vspace*{6pt}
Similarly, we use projection convolutions to map tokens on $\SOthree{}$ back to spherical signals:
\begin{bdefinition}[Projection convolution $\SOthree{} \to \Stwo{}$]
\label{def:proj_conv}
Let $f : \SOthree{} \rightarrow \mathbb{R}^c$ be a signal over the \SOthree{} group and
$\psi_\downarrow : \Stwo{} \rightarrow \mathbb{R}^{c_{\text{out}} \times c}$ a filter.
The \emph{projection convolution} of $f$ with $\psi_\downarrow$ is the map
$\psi_\downarrow \star f : \Stwo{} \to \mathbb{R}^{c_{
    \text{out}
}}$ given by
\begin{equation}
  (\psi_\downarrow \star f)(s) \ =\ \int_{\SOthree{}} \psi_\downarrow\big(R^{-1}s\big)\, f(R) \,\mathrm{d}R
  \qquad\textup{for any }\ \ s \in \Stwo{},
  \label{eq:proj-conv}
\end{equation}
\end{bdefinition}

\vspace*{6pt}
We define $\SOthree{}$-RoPE as acting via unitary representations of the rotation group.
For completeness, the following definition establishes what it means for a representation to be unitary:
\begin{bdefinition}[Unitary \SOthree{} representation]
    \label{def:unitary_so3_rep}
    We call $\pi : \SOthree{} \rightarrow \mathbb{R}^{c\times c}$
    a \emph{unitary group representation} if:
    
    \begin{enumerate}[label=\textup{(\Roman*)}]
        \item \label{def:unitary}
        $\pi(U)^\top\, \pi(U)\,=\, I$\quad\quad\ 
        for all $U \in \SOthree{}$ \hfill (unitary);

        \item \label{def:homomorphism}
        $\pi(U)\pi(V) = \pi(UV)$\quad
        for all $U,V \in \SOthree{}$ \hfill (group homomorphism).
    \end{enumerate}
    where $I\in\mathbb{R}^{c \times c}$ is the identity matrix.
\end{bdefinition}

\section{Proofs}\label{app:proofs}

\begin{proof}[Proof of Theorem~\ref{thm:rope_equivariant}]
Let $q,k,v : \SOthree{} \to \mathbb{R}^c$ be the query, key and value functions, and let $\pi : \SOthree{} \to \mathbb{R}^{c \times c}$ be a unitary representation (Def.~\ref{def:unitary_so3_rep}).

We first record two elementary properties of unitary representations. The identity element $e \in \SOthree{}$ is represented by the identity matrix $I\in \mathbb{R}^{c\times c}$:
\begin{equation}
\label{eqn:group_identity}
\begin{aligned}
    & \ \ \ \ \ \ \ \ \ \ \ \pi(U)\pi(e)\  \overset{\ref{def:homomorphism}}{=} \pi(U e) =\  \pi(U) \\
    \Rightarrow\  & \pi(U)^\top \pi(U) \pi(e) \ =\  \pi(U)^\top \pi(U)\\
    \overset{\ref{def:unitary}}{=}\ & \ \ \ \ \ \ \ \ \ \ \ \ \ \ \ \ \ \ \ \  \pi(e)\  = \ I
\end{aligned}
\end{equation}

Consequently, the inverse element $U^{-1}$ is represented by the inverse matrix $\pi(U)^{-1}$:

\begin{equation}
\label{eqn:group_inverse}
\begin{aligned}
    &\pi(U^{-1})\pi(U)\  \overset{\ref{def:homomorphism}}{=}\pi(U^{-1}U) \ \overset{\ref{eqn:group_identity}}{=}\  \pi(e) = I \\
    \Rightarrow \ & \pi(U^{-1}) = \pi(U)^{-1}
\end{aligned}
\end{equation}

Together, unitarity~\ref{def:unitary} and the composition property~\ref{def:homomorphism} give the main ingredient of the proof: the product $\pi(U)^\top \pi(V)$ depends on $U,V \in \SOthree{}$ only through their relative rotation $U^{-1}V$,
\begin{equation}
\label{eqn:relative_pi}
    \pi(U)^\top\, \pi(V)
    \ \overset{\ref{def:unitary}}{=}\ \pi(U)^{-1}\, \pi(V)
    \ \overset{\ref{eqn:group_inverse}}{=}\pi(U^{-1})\, \pi(V) \ \overset{\ref{def:homomorphism}}{=}\ \pi(U^{-1}V)
\end{equation}
In particular, the product is unchanged when both $U$ and $V$ are rotated by the same $R^{-1} \in \SOthree{}$:
\begin{equation}
    \pi\big((R^{-1}U)^{-1} \, R^{-1}V\big) = \pi(U^{-1} RR^{-1}\ V)= \pi(U^{-1} V)
\end{equation}

With this, the embedded attention score $A_{\text{RoPE}}[q,k] := A\big[\mathbf{\Upsilon}[q],\mathbf{\Upsilon}[k]\big]$ is rotation-equivariant: for all $R\in \SOthree{}$ and all $U,V \in \SOthree{}$,
\begin{equation}
\label{eqn:attn_score_eqi}
\begin{aligned}
 & A_{\text{RoPE}}[R \rhd q,\, R \rhd k](U,V)\\
 &\overset{\text{Def.~\ref{def:rope}}}{=}\ \mathbf{\Upsilon}[R \rhd q](U)^\top\, \mathbf{\Upsilon}[R \rhd k](V) \\
 &\ \ \ =\, \, \ \ \ q(R^{-1}U)^\top\, \pi(U)^\top\pi(V)\ k(R^{-1}V) \\
 &\ \, \overset{(\ref{eqn:relative_pi})}{=}\ \ \ q(R^{-1}U)^\top\, \pi(U^{-1}V)\, k(R^{-1}V) \\
 &\ \ \ =\, \, \ \ \ q(R^{-1}U)^\top\, \pi(U^{-1}R\, R^{-1}V)\  k(R^{-1}V) \\
 &\ \, \overset{(\ref{eqn:relative_pi})}{=}\ \ \ q(R^{-1}U)^\top\, \pi(R^{-1}U)^\top\, \pi(R^{-1}V)\, k(R^{-1}V) \\
 &\overset{\text{Def.~\ref{def:rope}}}{=}\ A\big[\mathbf{\Upsilon}[q],\mathbf{\Upsilon}[k]\big](R^{-1}U,R^{-1}V)\\
 &\ \ \ =\ \ \ \ A_{\text{RoPE}}[q,k](R^{-1}U,R^{-1}V).
\end{aligned}
\end{equation}

The softmax inherits this equivariance. Its denominator integrates over all of \SOthree{}, so we may substitute $D := R^{-1}Q$ without changing the integral.

\begin{equation}
\label{eqn:softmax_eqi}
\begin{aligned}
\operatorname{softmax}\big[
& A_{\text{RoPE}}[R \rhd q,\, R \rhd k]
\big](U,V) \\
&\ \ \ \ \ =\ \ \ \ \ \ \,
\frac{
e^{A_{\text{RoPE}}[R \rhd q,\, R \rhd k](U,V)}
}{
\int_{\SOthree{}} e^{A_{\text{RoPE}}[R \rhd q,\, R \rhd k](U,Q)} \,\mathrm{d}Q
}\\
&\ \ \ \ \overset{(\ref{eqn:attn_score_eqi})}{=}\ \ \ \ \ \ 
\frac{
e^{A_{\text{RoPE}}[q,k](R^{-1}U,R^{-1}V)}
}{
\int_{\SOthree{}} e^{A_{\text{RoPE}}[q,k](R^{-1}U,R^{-1}Q)} \,\mathrm{d}Q
}\\
&\overset{D:=R^{-1}Q}{=}\ 
\frac{
e^{A_{\text{RoPE}}[q,k](R^{-1}U,R^{-1}V)}
}{
\int_{\SOthree{}} e^{A_{\text{RoPE}}[q,k](R^{-1}U,D)} \,\mathrm{d}D
}\\
&\ \ \ \ \ \ \,=\ \ \ \ \ \ \,
\operatorname{softmax}\big[A_{\text{RoPE}}[q,k]\big](R^{-1}U,R^{-1}V)
\end{aligned}
\end{equation}

Similar, the RoPE-embedded attention is \SOthree{}-equivariant in its argument $U \in \SOthree{}$:

\begin{equation}
\begin{aligned}
\operatorname{attn}\big[
&\mathbf{\Upsilon}[R \rhd q],\, \mathbf{\Upsilon}[R \rhd k],\, R \rhd v
\big](U)\\
&\ \ \ \, \overset{\text{Def.~\ref{def:cont_attn}}}{=}\ \ \ \ 
\int_{\SOthree{}}
\operatorname{softmax}\big[
A_{\text{RoPE}}[R \rhd q,\, R \rhd k]
\big](U,V)\, (R \rhd v)(V) \,\mathrm{d}V\\
&\ \ \ \ \, \overset{(\ref{eqn:softmax_eqi})}{=}\ \ \ \ \ \  
\int_{\SOthree{}}
\operatorname{softmax}\big[
A_{\text{RoPE}}[q,k]
\big](R^{-1}U,R^{-1}V)\, v(R^{-1}V) \,\mathrm{d}V\\
&\ \overset{D:=R^{-1}V}{=}\ 
\int_{\SOthree{}}
\operatorname{softmax}\big[
A_{\text{RoPE}}[q,k]
\big](R^{-1}U,D)\, v(D) \,\mathrm{d}D\\
&\ \ \ \ \ \overset{\text{Def.~\ref{def:cont_attn}}}{=}\ \ \ \ \ \ \ \ 
\operatorname{attn}\big[\mathbf{\Upsilon}[q],\, \mathbf{\Upsilon}[k],\, v\big](R^{-1}U)\\
&\ \ \ \ \ \ \ \,=\big(R \rhd \operatorname{attn}\big[\mathbf{\Upsilon}[q],\, \mathbf{\Upsilon}[k],\, v\big]\big)(U)
\end{aligned}
\end{equation}

\end{proof}

\end{document}